\documentclass{article}
\usepackage{spconf,amsmath,graphicx}
\usepackage{booktabs}
\usepackage{graphicx}
\usepackage{subfig}
\usepackage{xcolor}
\usepackage{times}
\usepackage{latexsym}
\usepackage{tabularx}
\usepackage{soul}

\newcolumntype{b}{X}
\newcolumntype{s}{>{\hsize=.5\hsize}X}

\title{Towards Language Modelling in the Speech Domain Using Sub-word Linguistic Units}
\name{Anurag Katakkar; Alan W Black}
\address{Carnegie Mellon University;\\
        \{akatakka, awb\}@andrew.cmu.edu}
\begin{document}
%
\maketitle
\begin{abstract}
Language models (LMs) for text data
have been studied extensively
for their usefulness in
language generation and other
downstream tasks. 
However, language modelling purely in 
the speech domain is still a relatively
unexplored topic, with traditional speech
LMs often depending on auxiliary text LMs
for learning distributional aspects
of the language. For the English language,
these LMs treat words 
as atomic units, which
presents inherent challenges to language
modelling in the speech domain.
In this paper, we propose a novel
LSTM-based
generative speech
LM that is inspired by the
CBOW model and built on linguistic
units including syllables and phonemes.
This offers better acoustic consistency 
across utterances in the dataset, 
as opposed to single melspectrogram
frames, or whole words. 
With a limited dataset, orders of 
magnitude smaller than 
that required by
contemporary generative models,
our model
closely approximates babbling speech.
We show the effect of training 
with auxiliary text LMs, multitask learning
objectives, and auxiliary articulatory
features. Through our experiments, 
we also highlight some well known,
but poorly documented challenges in training 
generative speech LMs, including the mismatch
between the supervised learning objective
with which these models are trained
such as Mean Squared Error (MSE), and
the true objective, which is speech quality.
Our experiments provide an early indication
that while validation loss and Mel Cepstral
Distortion (MCD) are not strongly
correlated with generated speech quality,
traditional text language modelling 
metrics like perplexity and 
next-token-prediction accuracy might be.
\end{abstract}
\section{Introduction}
\begin{figure*}[htp]
  \centering \includegraphics[width=15cm, height=8cm]{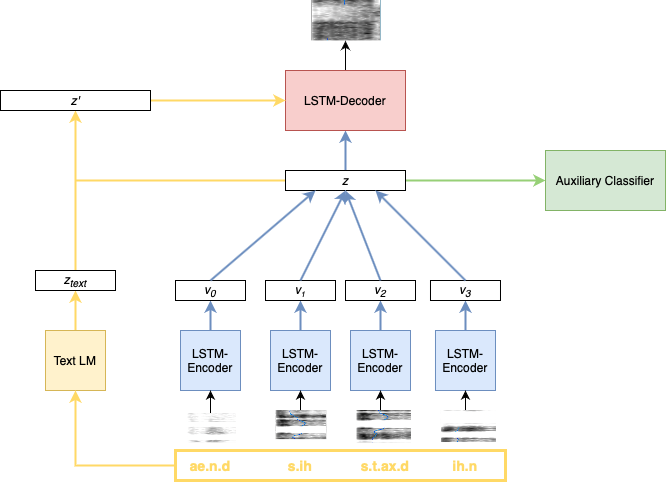}
  \caption{A schematic diagram of our model(s). The blue arrows indicate the speech-only LM, yellow arrows the text LM augmented model, and the green arrow shows the multitask learning setting with the auxiliary classifier. Boxes of the same color share parameters.}
  \label{fig:model}
\end{figure*}

From traditional n-gram models
to more recent LSTM and transformer 
based neural models, language models (LMs)
in the text domain have been explored
extensively \cite{Mikolov2010RecurrentNN, Mikolov2013EfficientEO, Devlin2019BERTPO, Brown2020LanguageMA}.
Their performance on language
modelling tasks as measured by token prediction
accuracy and perplexity, and on downstream
natural language processing tasks such as
question answering, natural language 
inference, and sentiment analysis among 
others, have been the subject
of several studies 
including some that have
even helped develop an understanding of
how well these models scale with the
amount of training data, and the 
number of model parameters \cite{Kaplan2020ScalingLF}. 
Studies that probe text neural
language models for real world knowledge
have established methods for understanding
what information is stored in
these LMs and at what point during 
training they begin to know it 
\cite{Jiang2020HowCW, Jiang2021HowCW, Shin2020ElicitingKF}.
With GPT-3 \cite{Brown2020LanguageMA}, 
generative language models have crossed
the 100 billion parameter mark and can
produce fluent natural language text
that is difficult to distinguish from
human generated text.
In general, it may be said that contemporary
text LMs are both, good representation learners, and language generators.

Language models in the speech domain,
however, have been explored relatively less.
Models such as wav2vec  \cite{Schneider2019wav2vecUP} and 
Mockingjay \cite{Liu2020MockingjayUS}
show state-of-the-art performance
on word- and phoneme-error-rate, and 
other downstream tasks such as speaker
identification and sentiment analysis.
While these are good representation
learners, they are not speech synthesizers.
On the other hand, Text-To-Speech (TTS)
models such as Tacotron \cite{Wang2017TacotronTE},
are just that,
and therefore not pure speech language models.
``Textless NLP" models such as 
Slow AutoEncoders (SlowAEs) \cite{Dieleman2021VariablerateDR} and 
the Generative Spoken Language Model
(GSLM) \cite{Lakhotia2021GenerativeSL} are amongst the first that
attempt to build language models 
in the speech domain directly and
purely from raw speech (waveform).
However, these models use hundreds or
thousands of hours of speech data, and 
models whose parameters number
in the hundreds of millions, or billions -
both several orders of magnitude
greater than our proposed method.
Further, unlike our
method that uses linguistic units,
they use entire 
utterances of raw speech as input, 
limiting applications 
to single language systems.

Generative speech models use the raw
speech waveform, or melspectrograms 
generated from raw speech as their
input, and a single frame as the
computational unit of speech. Single
melspectrogram frames contain
about 5-10ms of speech or less, which 
is smaller than the average length
of a phoneme in English. This short
length prevents a single frame from 
containing much inherent linguistic 
structure, and leads to a large
variation in frames across different
parts of the entire melspectrogram.
Instead, using linguistic units such as 
phonemes
and syllables, offers more consistency
in structure across the utterance. 
As a first
step towards linguistically grounded
speech LMs, we propose a simple
LSTM based encoder-decoder model
that babbles.
In our 
experiments, we show that using
auxiliary information such as
articulatory linguistic features, 
or embeddings from text LMs
built on the same linguistic units
(phonemes or syllables) can improve
synthesis quality considerably. We also 
find that multi-task
learning objectives that encourage 
the model to capture these linguistic
features help improve synthesis quality.

Throughout the paper, we discuss several 
well known, but poorly documented challenges
of training speech synthesis models such as 
the disagreement between generated speech
quality and the supervised learning 
objective, which consequently make
the use of learning curves unreliable
in decisions concerning duration of
training and objectively identifying
better models.

\section{Related Work}
\begin{table*}[t!]
\centering
\small
\begin{tabularx}{\textwidth}{sbb}
    \toprule
    Dataset & Original Text & Syllabified Text \\
    \midrule
     LJSpeech & Printing in the only sense with which we are at present \dots & $\langle p.r.ih.n\rangle$$\langle t.ih.n.g\rangle$$\langle ih.n\rangle$\hl{$\langle dh.ax\rangle$}$\langle ow.n\rangle$$\langle l.iy\rangle$$\langle s.eh.n.s\rangle$ $\langle w.ih.dh\rangle$$\langle w.ih.ch\rangle$$\langle w.iy\rangle$$\langle aa.r\rangle$$\langle.ae.t \rangle$$\langle p.r.eh\rangle$
     $\langle z.ax.n.t\rangle$ \dots
     \\
     LJSpeech & To the various wards their friends occupied \dots & 
     $\langle t.ax\rangle$\hl{$\langle dh.ax\rangle$}$\langle v.eh\rangle$$\langle r.iy\rangle$$\langle ax.s\rangle$\hl{$\langle w.ao.r.d.z\rangle$}
$\langle dh.eh.r\rangle$\hl{$\langle f.r.eh.n.d.z\rangle$}$\langle aa\rangle$$\langle k.y.ax\rangle$$\langle p.ay.d\rangle$ \dots
    \\
\bottomrule
\end{tabularx}
\caption{Examples of utterances from the LJSpeech dataset with original text and syllabified text. For the purpose of readability, angular brackets and periods are used to separate individual syllables and phonemes respectively. Highlighted syllables are removed after preprocessing either because they are too long (longer than 250ms) or are stop words.}
\label{tab:syllabified-examples}
\end{table*}

\textbf{Language Modelling in Text} 
In their most
simple definition, language models are
statistical models that can assign 
a probability score to a given sequence
of tokens
\cite{Jurafsky2000SpeechAL}. Further,
and particularly important to our work,
language models that are trained in 
a purely auto-regressive manner
(with left context only) can be used 
to generate language.
Early recurrent
neural network language models were already
quite successful in 
outperforming traditional
n-gram based language models
\cite{Mikolov2010RecurrentNN}
measured by objective metrics such
as perplexity and accuracy of 
next-token prediction. 
The skip-gram and CBOW models presented in
\cite{Mikolov2013EfficientEO} serve as the
baseline architecture that we replicate in
the speech domain. 
Transformer based language models
such as BERT \cite{Devlin2019BERTPO},
and RoBERTa \cite{Liu2019RoBERTaAR}
make use of bi-directional encodings
and cannot be considered true language 
models in this sense. The use
of bi-directional context inhibits their
use in generating language.

\textbf{Speech Synthesis} Though Variational
Autoencoders (VAE) and Generative Adversarial
Networks (GANs) have been used in speech
synthesis \cite{Binkowski2020HighFS}, 
they are not language models
in that they do not possess any of the 
aforementioned LM properties - specifically, 
being conditioned on the left context, and being
able to predict the next token in speech. 
Text-to-Speech models are closer to being 
language models, but are not pure speech
language models. Modern TTS models are also
not strictly left-to-right conditioned because
of their use of Transformer
\cite{Vaswani2017AttentionIA} 
style bi-directional
architectures. Several TTS models now use
auxiliary features from text LMs
to enhance the quality of generated speech.
In \cite{Hayashi2019PreTrainedTE}
the use of auxiliary embeddings from
a BERT text language model is reported
to produce significant improvement in
the quality of generated speech.
The cause for this improvement and the
precise information these auxiliary features
contain is not investigated.
Continuing in this line of research,
we experiment
with auxiliary features from two text LMs
- including one non-Transformer model -
and study their effect on the quality of
synthesised speech. Our findings 
indicate that the accuracy and perplexity
of these auxiliary models plays a
significant role in the quality of speech
produced.

\textbf{The use of Sub-word Linguistic Units} The choice of the fundamental 
unit for language modelling
in speech is not as straightforward
as it is in the text domain
where languages like English are
generally tokenized on whitespace.
In speech, sub-word models are 
more closely investigated in languages
such as Mandarin 
\cite{Zhou2018SyllableBasedSS}
although applications are mostly limited
to ASR, and do not include synthesis.
\cite{Livescu2012SubwordMF} argues for
the use of sub-word models 
in Automatic Speech Recognition
because of their advanced 
ability to handle
large vocabulary corpora with many
low-frequency tokens. 
Sub-word units also provide a
more consistent acoustic representation
in comparison to entire words
or single melspectrogram frames.
Particularly, the use of phonemes allows us
to work with a limited vocabulary, with only
39 phonemes in the English CMU Phoneme 
Dictionary. The use of syllables, which
in our definition always contain a
central vowel, also offer consistency
in the acoustic domain. 

\textbf{Articulatory features} 
Another challenge in speech processing
is the great variation in the manner of
articulation of phones, which is also
often context dependent.

Previous work has argued for representing 
speech as "multiple parallel streams of
information". Each stream provides information
about different articulatory features of the
speech segment such as being voiced or unvoiced.
\cite{Mortensen2016PanPhonAR} presents a novel
toolkit for representing IPA speech
segments with 22
articulatory features. In their experiments, 
the authors report the use of these articulatory
features in conjunction with word-vector based 
methods achieves superior 
performance on NLP tasks such as Named Entity
Recognition, as compared to word-vector methods
alone. 
The Panphon toolkit proposed in \cite{Mortensen2016PanPhonAR}
converts IPA speech segments into 22
articulatory features, and significantly
improves performance over word-vector
based methods, when used as auxiliary
features in conjunction with thse
dense-vector representations.
In our experiments, we make use of these
articulatory features as auxiliary inputs to 
our speech LM, as well as in a multitask 
setting in which the latent representation
from our LSTM-encoder
is used to predict these features. This is 
discussed in further detail in section 4.

\textbf{Speech quality assessment} A well known
problem in speech synthesis is the lack of 
objective methods of evaluating the quality 
of synthesised speech. From observations made
by training several models (with reasonable 
exploration of the hyperparameter space) it is
also clear that there is no strong correlation 
between the optimization objective used to train
speech synthesis models and the quality of 
synthesised speech. MOS is a widely used 
metric but given its subjective nature it is
impossible to make direct comparisons between
results from different studies without re-implementing
the proposed models, and conducting a fresh round
of human evaluation. We make use of Mel-Cepstral-Distortion (MCD) in our 
experiments and also evaluate the latent 
representations of our model on downstream 
language modelling tasks such as next token prediction
and report the correlation between model performance
on these tasks and MOS. 

\section{Dataset}
We use the LJSpeech dataset for 
all models and experiments in the 
speech domain. This dataset contains
24 hours of read English speech from 
audio-books and is part of the LibriVox 
project. For phoneme-level LMs, each
utterance is split into constituent 
phonemes according to alignments
generated by [todo].
For syllable-level
LMs, we first define a syllable as
a unit that contains at least a vowel, 
and optionally a pre-vowel and post-vowel made up of one or more consonants.
Syllable alignments are generated by
[todo - 
Syllables longer than 250ms, phonemes 
longer than 150ms, and all silences are 
removed from the dataset. This ultimately
leaves us with approximately 18 hours 
of speech.

Note that we convert all of our 
speech data into melspectrogram
representations of the waveform.
For articulatory linguistic features
for syllables we use the Panphon toolkit
\cite{Mortensen2016PanPhonAR}
that maps phonemes to 22 subsegmental
articulatory features. 
When the pre- or post-vowel contain more
than one consonant, we combine their
representations using max-pooling to obtain
a single feature vector for the pre-vowel, 
and the post-vowel. 
Finally, we concatenate the resulting 
feature vectors of the pre-vowel, vowel, 
and post-vowel to obtain a single 
66-dimensional vector.

For our text LMs that are used to 
provide auxiliary information to the 
speech LM, we use a subset of the English
Wikipedia dataset that contains 
approximately 30 million tokens. Note
that since we want to use the same 
units for the speech and text LMs, 
this text is also syllabified, or
converted to phonemes, in a similar
manner to the utterances in the
LJSpeech dataset. Table \ref{tab:syllabified-examples} contains examples of syllabified text from the LJSpeech dataset.

\section{Experiments}
\begin{figure}
  \includegraphics[width=\linewidth]{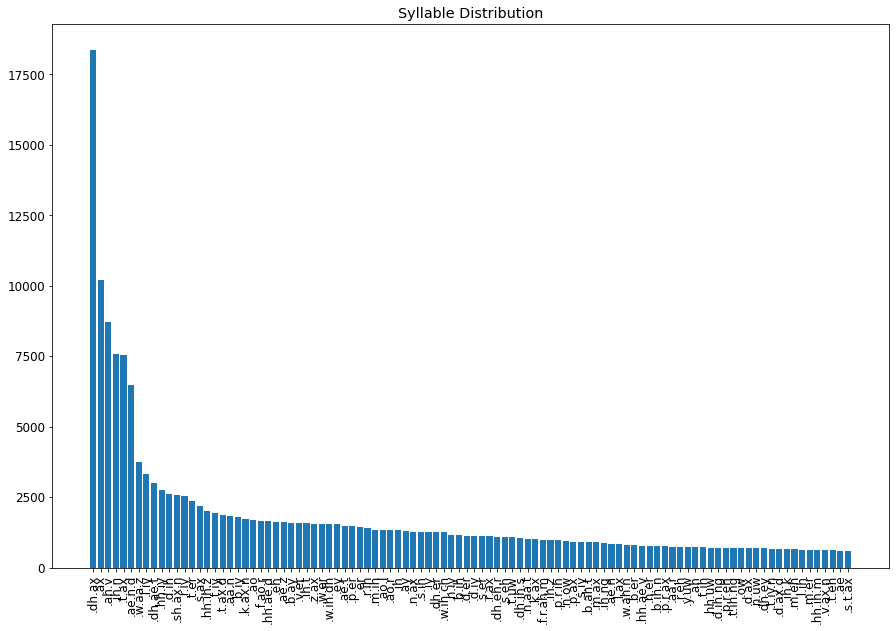}
  \caption{Distribution of top 100 syllables in the preprocessed LJSpeech dataset.}
  \label{fig:ljspeech-syllables}
\end{figure}

A simplified diagram of our model(s) is shown in Figure ~\ref{fig:model}. Below, we describe the details of all our models and experiments.

\textbf{Pure Speech Language Model}  
As our first approximation, we 
construct a simple LSTM-based
 encoder-decoder 
language model (Synthesis-only model in 
Table ~\ref{tab:results1})
inspired by the CBOW model \cite{Mikolov2013EfficientEO}.
For this model to be a true language
model, we only provide it with the left
context, and frame the problem to 
predict the next unit given this context.
In our experiments, we find 4 units 
(syllables or phonemes) of 
context to be optimal.
The encoder LSTM first
extracts 256-dimensional latent
representations from the melspectrogram
of each of the 4
context units independently, $v_i$. These
latent representations are then 
concatenated into a single 1024-dimensional
vector ($z$) and used as an input to the
decoder LSTM
to predict the melspectrogram frames 
of the next syllable. The decoder thus
models the auto-regressive LM objective
$p(y | z)$, where $y$ is the next unit
to be predicted given the context $z$.
Loss is computed
as the Mean Squared Error between the
predicted and the ground truth melspectrogram.

\textbf{Multitask Learning Objectives} 
In order to encourage the model to
produce more diverse sounding units
without using more speech data, we
use an auxiliary objective to predict
the Panphon representation of the
next unit and train in a Multitask
Learning setting.
Concretely, the 1024-dimensional latent
representation is additionally used as
input to an auxiliary
classifier that predicts
the 66-dimensional Panphon vector of 
the next syllable. We treat this as 
binary classification , and the loss
is computed as an average of binary
cross entropy over 66 independent 
variables.
 In Table 
~\ref{tab:results1} these models are
listed as MTL (Panphon).

\textbf{Auxiliary Features}
This method is already widely used in
several state of the art Text-to-Speech
systems \cite{Hayashi2019PreTrainedTE}.
For this, we first train two types of 
text language models on our Wikipedia
text dataset. Note that the text in this
dataset has been preprocessed identically to 
the transcripts of the LJSpeech dataset
for the respective sub-word linguistic 
unit type.
We experiment with two types of 
text language models - an LSTM language model
whose embedding matrix is initialised using a 
word2vec model trained on the same text data, 
and transformer based RoBERTa model 
\cite{Liu2019RoBERTaAR}. Note that
since our vocabulary no longer contains English
words, and instead is comprised of
phonemes or syllables, we train the 
RoBERTa model and its 
corresponding tokenizer from scratch.
The representations from these models, $z_{text}$
are then concatenated to the
latent representation $z$ extracted by 
the speech LSTM-encoder and
fed into the decoder. We use a fixed 
embedding size of 768 for the text LMs, 
therefore, the new latent representation
$z'$ is a $1792$ dimensional vector. 
The weights of the text language models are 
frozen after training and are not updated
when training the speech language model.

Finally, since there is no comparable baseline
method, to the best of our knowledge, that
performs speech language modelling
similar to our method, we suggest the use of a
``top-line" instead. This is meant to represent
the highest quality of speech that a sub-word
linguistic unit based speech language model
can generate when given the ground-truth
linguistic information about the next 
unit to be predicted. We model this by
providing the Panphon representation vector
as auxiliary information and concatenating
it to the extracted latent representation
of the context speech units, $z$. 
Naturally,
such a setting is not possible in a real
world scenario where the next unit is unknown
and to be predicted. However, we find it a
good demonstration of the potential of 
using articulatory features of sub-word units.
This model is listed as ``Auxilliary (Top-line)" in Table ~\ref{tab:results1}.



\section{Discussion and Analysis}
\begin{figure*}[htp]
\centering
\subfloat[]{
  \includegraphics[width=5cm]{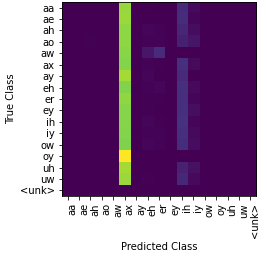}
}
\subfloat[]{
  \includegraphics[width=5cm]{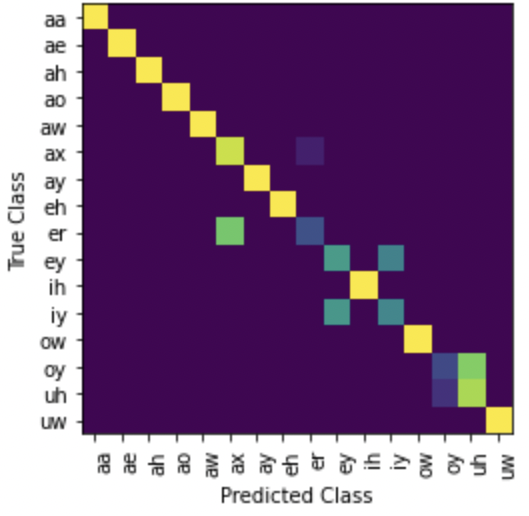}
}
\subfloat[]{
  \includegraphics[width=5cm]{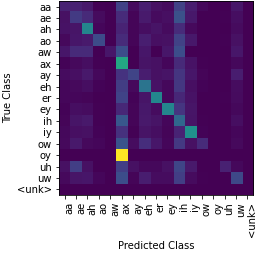}
}
\caption{Confusion Matrices of the post-hoc vowel classifier built from the Pure Speech LM latent representation, Panphon representation, and the Multitask Learning model
trained to predict Panphon features.}
\label{fig:confusion-matrices}
\end{figure*}

In this section we make several 
observations from the babbling
speech generated by our various models 
and provide insights into the observed
phenomena.

\emph{\textbf{Observation 1} Using the
syllabified LJSpeech dataset directly
results in a low diversity in the 
generated speech.}

\begin{figure*}[h!]
\centering
\subfloat[Reference]{
  \includegraphics[width=65mm]{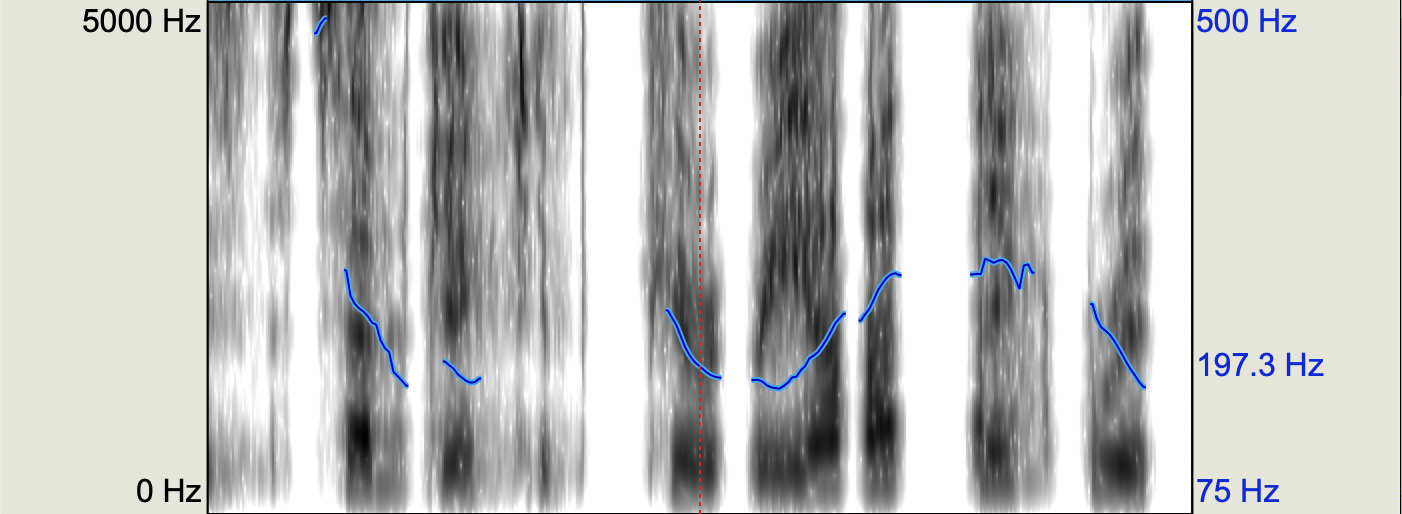}
}
\hspace{0mm}
\subfloat[Synthesis-only - 6 hours]{
  \includegraphics[width=65mm]{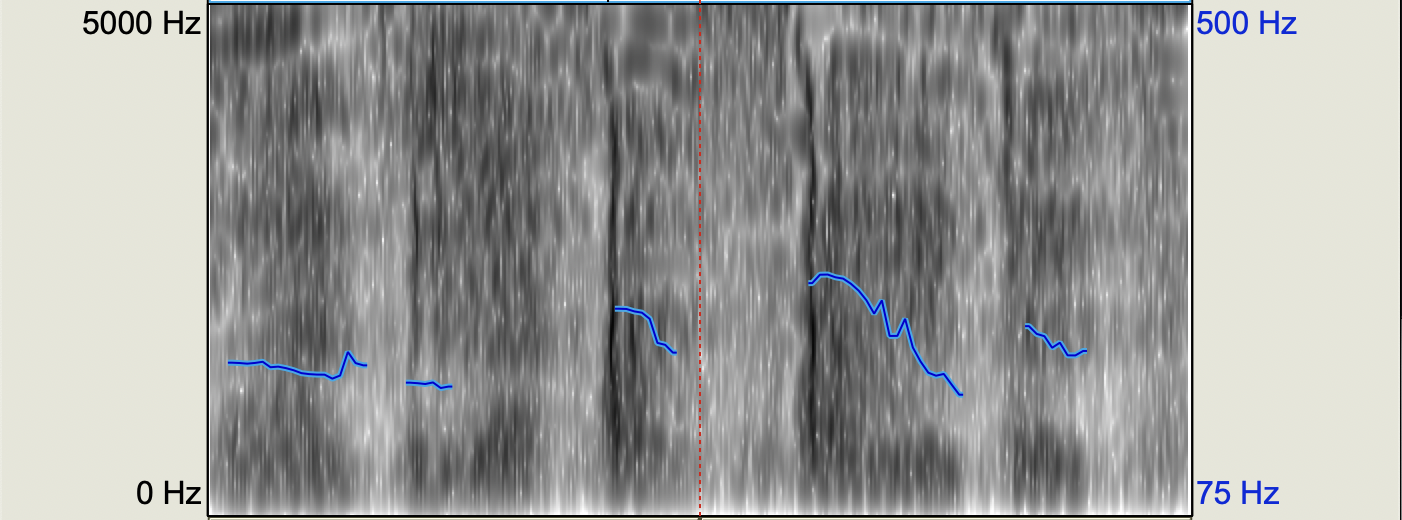}
}
\subfloat[Synthesis-only - 12 hours]{
  \includegraphics[width=65mm]{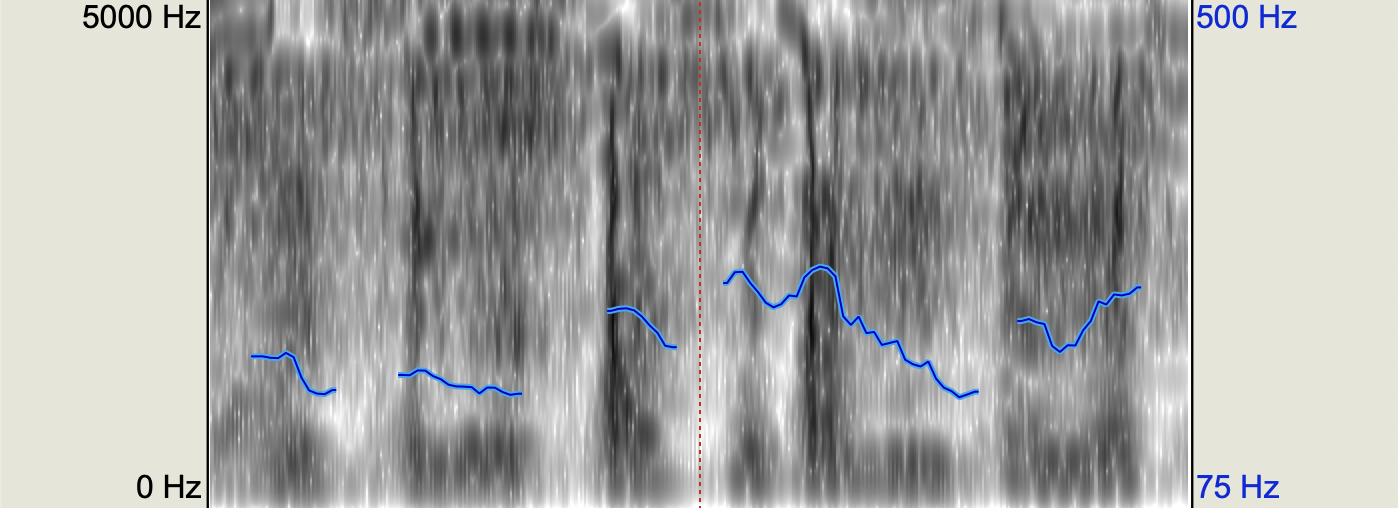}
}
\hspace{0mm}
\subfloat[MTL Panphon - 6 Hours]{   
  \includegraphics[width=65mm]{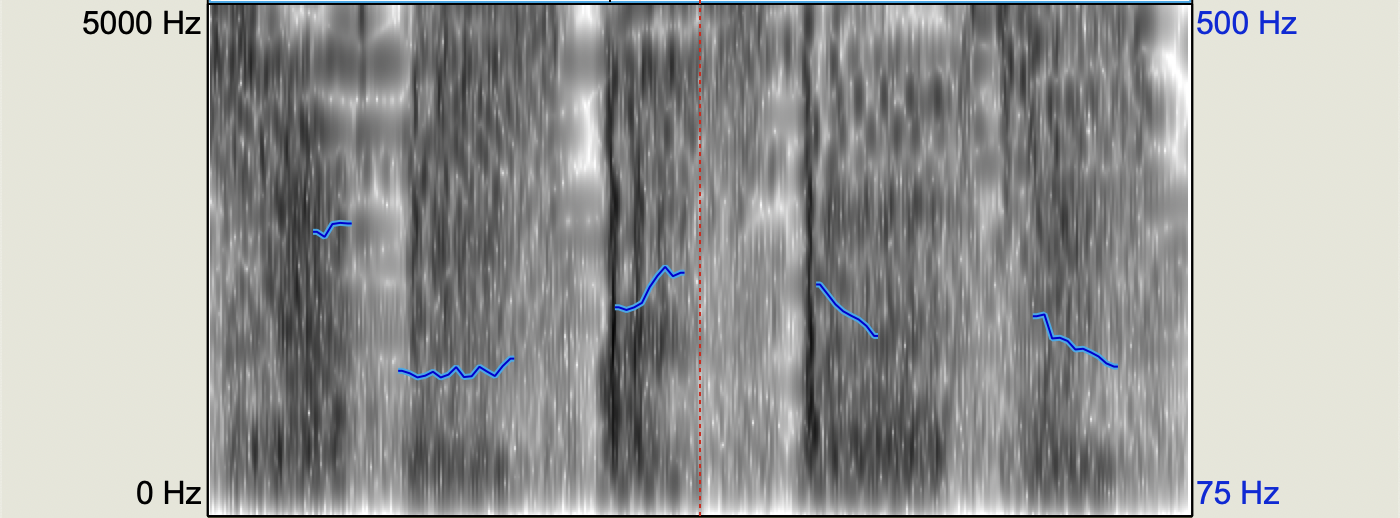}
}
\subfloat[MTL Panphon- 12 Hours]{
  \includegraphics[width=65mm]{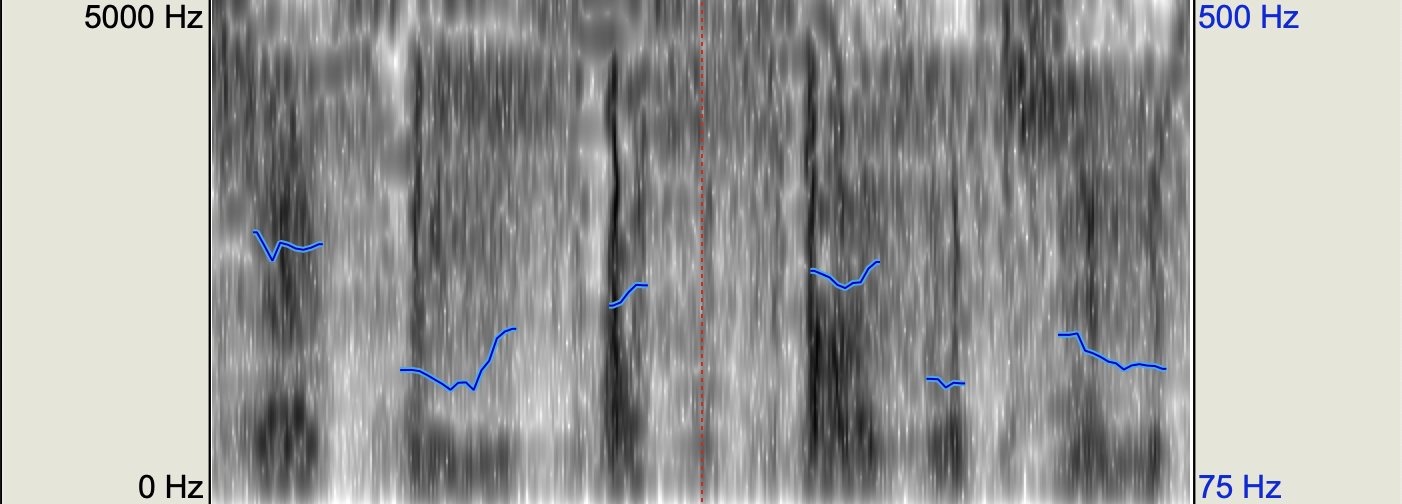}
}
\hspace{0mm}
\subfloat[Aux. LSTM LM - 6 Hours]{   
  \includegraphics[width=65mm]{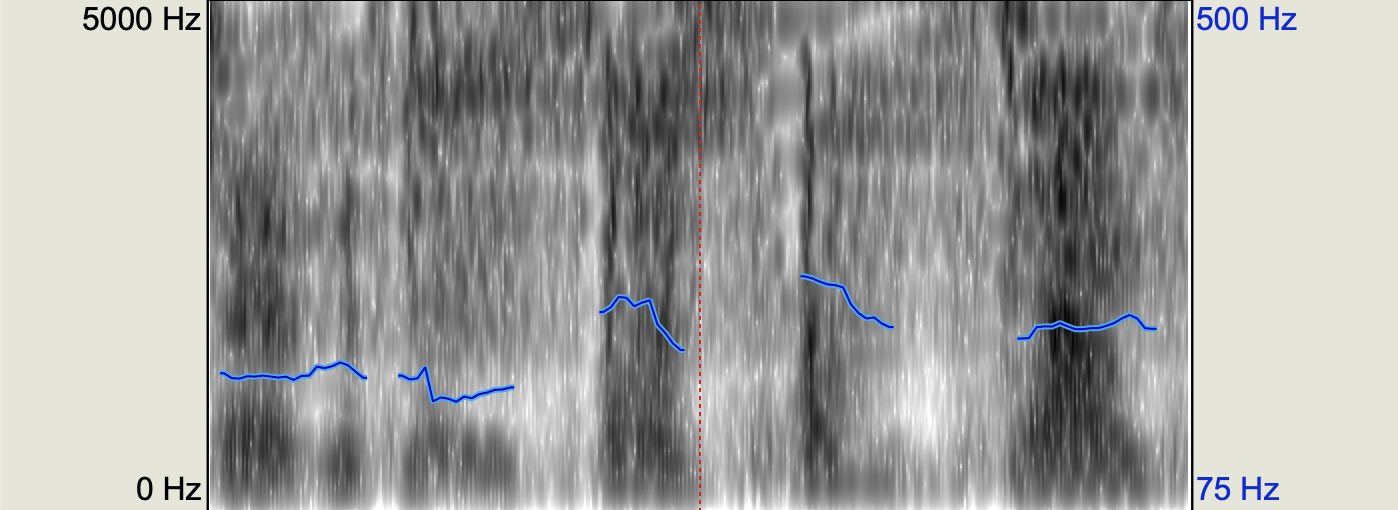}
}
\subfloat[Aux. LSTM LM - 12 Hours]{
  \includegraphics[width=65mm]{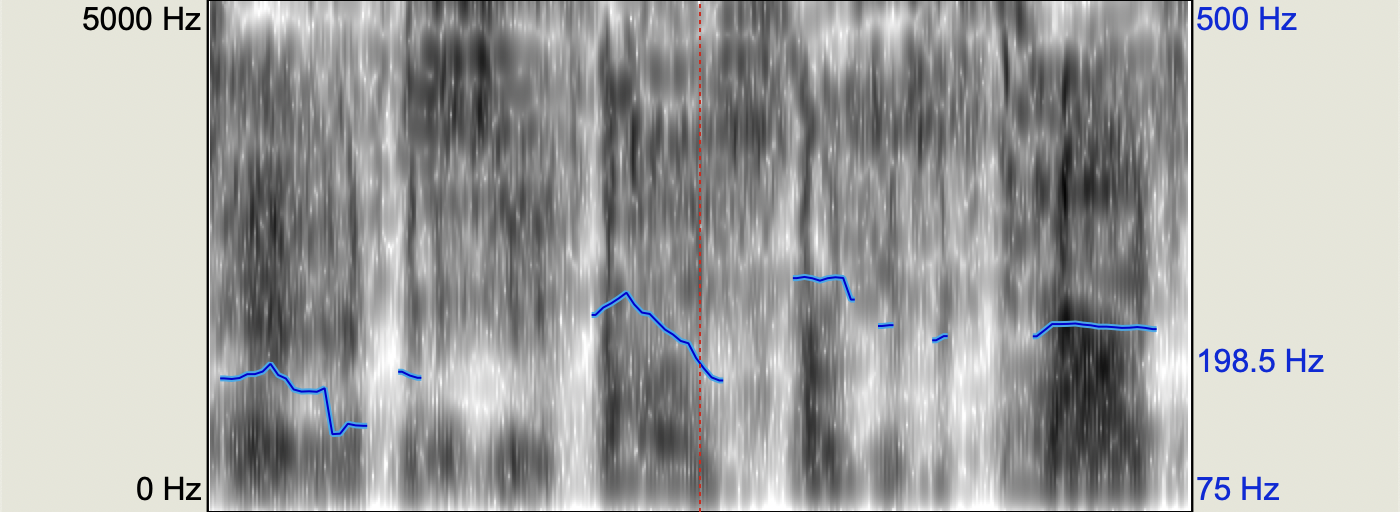}
}
\hspace{0mm}
\subfloat[Aux. RoBERTa LM - 6 Hours]{   
  \includegraphics[width=65mm]{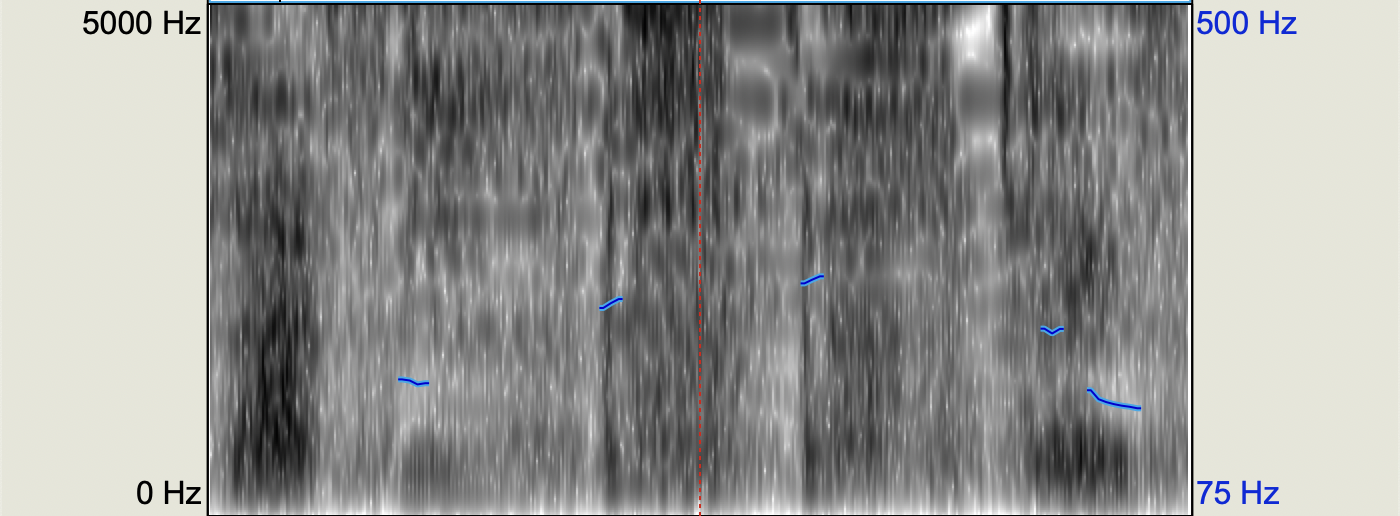}
}
\subfloat[Aux. RoBERTa LM - 12 Hours]{
  \includegraphics[width=65mm]{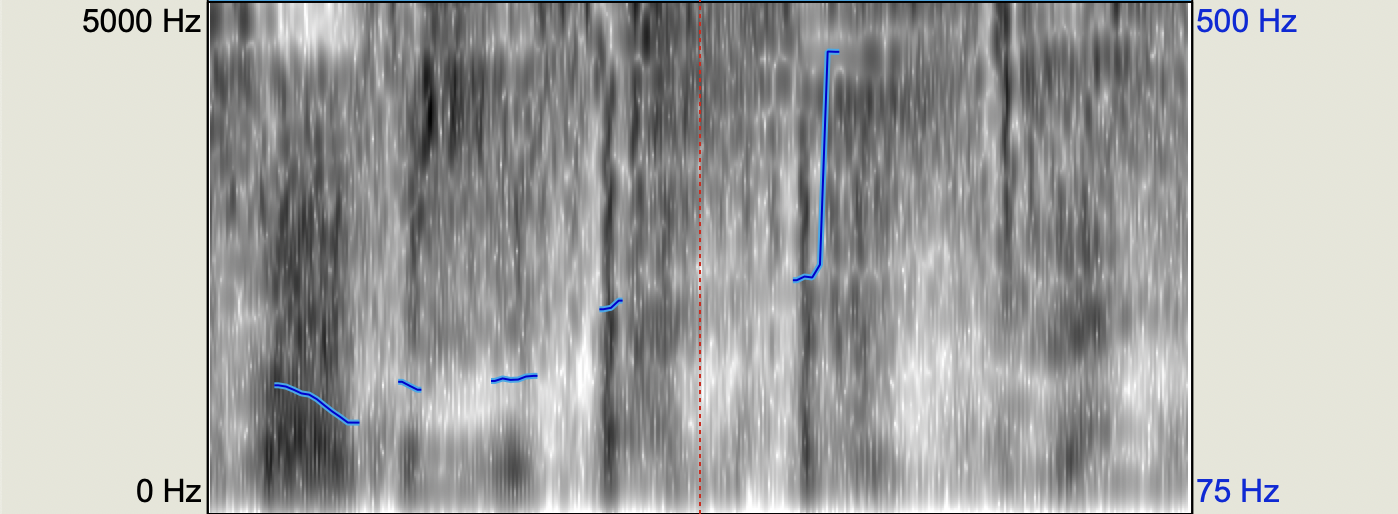}
}
\hspace{0mm}
\subfloat[Top-Line - 6 Hours]{   
  \includegraphics[width=65mm]{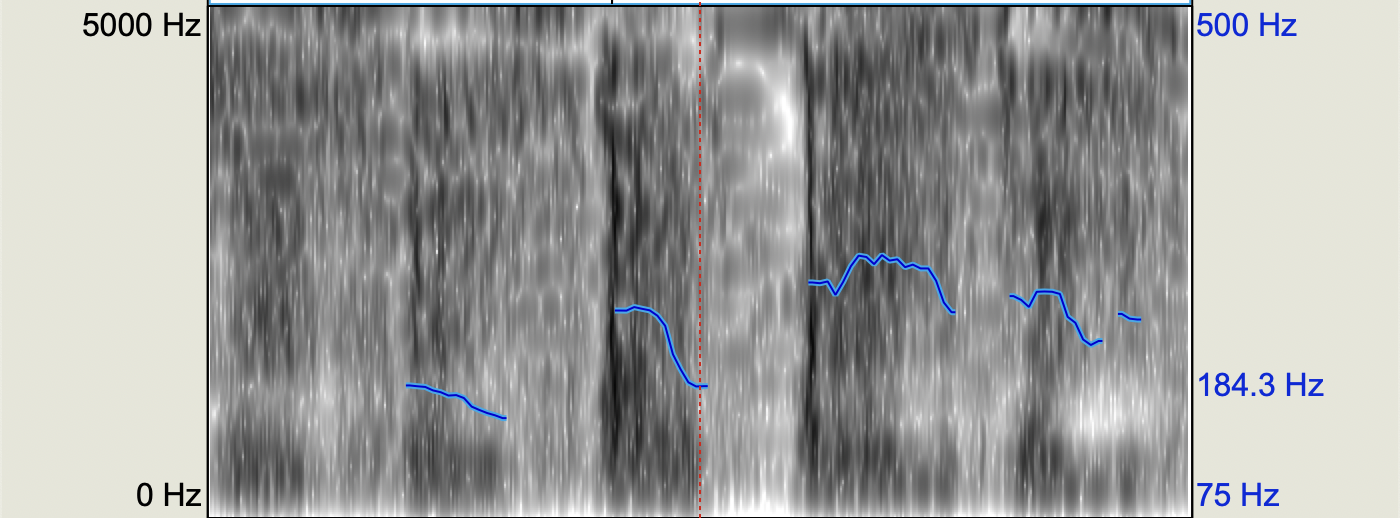}
}
\subfloat[Top-Line - 12 Hours]{
  \includegraphics[width=65mm]{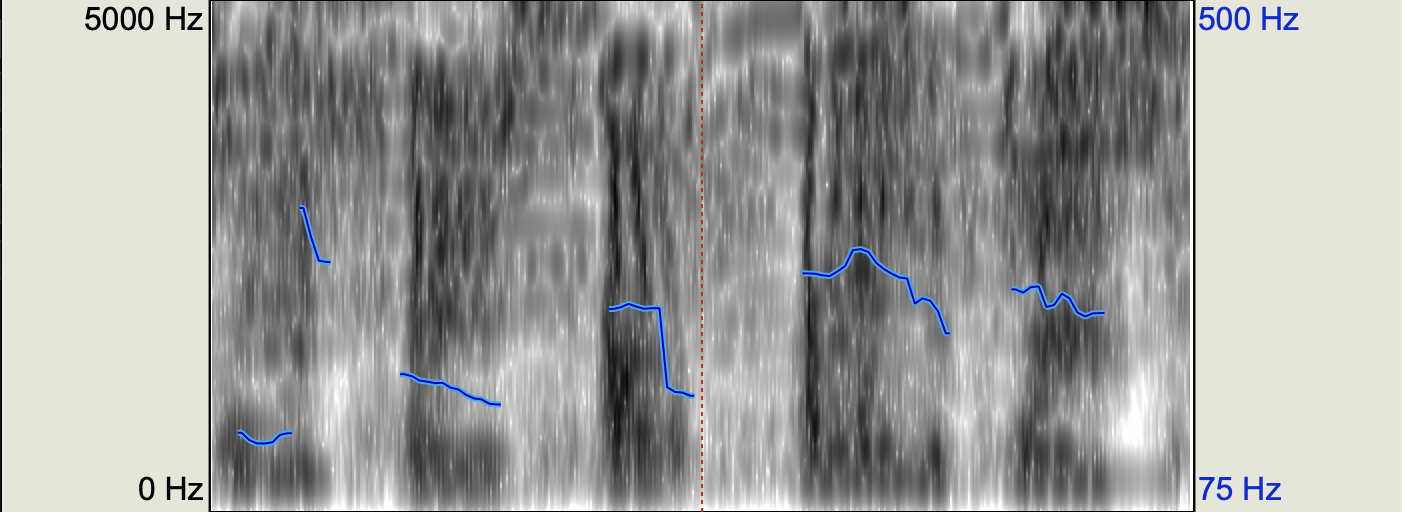}
}
\caption{Melspectrograms generated from the various models after 6 and 12 hours of training respectively. Clearly, as training progresses, more syllabic structure emerges in the spectrograms.}
\label{fig:spectrograms}
\end{figure*}

Our first approximation (Pure Speech LM)
used the LJSpeech dataset that was 
preprocessed only to remove pauses
and syllables or phonemes that were
exceptionally short or long.
However, the syllable distribution,
in this dataset, 
shown in Figure 
\ref{fig:ljspeech-syllables},
is highly 
skewed, with ``ax" and ``dhax" 
together making up about than 10\%
of the dataset which has approximately
$5000$ syllables. The generated
samples contain repetitive speech
that resembles the ``ax'' sound
and this repetition can be seen
clearly in the spectrograms in 
Figure \ref{fig:spectrograms}.

\emph{\textbf{Observation 2} The latent
representation of the Pure Speech LM lacks discriminative information
about the predicted token}

To investigate whether the latent
representation extracted by the 
LSTM-encoder contains information
about the next token to be predicted,
we build a post-hoc phoneme classifier.
Since the vowel is central to our
definition of the syllable, and 
since the number of phonemes (16) is
significantly smaller than the 
number of syllables in the dataset,
we hypothesise that phoneme 
classification is an ideal test 
for information contained in the
latent representation. For the pure 
speech language model, we observe
that this defaults to a majority
classifier as seen from the
confusion matrix in Figure \ref{fig:confusion-matrices} (a). 

\emph{\textbf{Observation 3} Panphon 
features contain important linguistic
information that is useful to
speech language modelling.}

When the same post-hoc classifier
is trained with Panphon representations
directly, accuracy improves greatly,
Figure \ref{fig:confusion-matrices} (b). This
confirms that the Panphon features
do contain important linguistic 
information. Guided by this observation
we select the prediction of Panphon 
features as an auxiliary task in our
multitask learning setting with the
hope that it encourages more linguistic
information to be captured in the 
latent representation that is fed into
the decoder. Indeed, as seen from 
Figure
\ref{fig:confusion-matrices} (c)
there is an improvement in 
vowel prediction performance. This
improvement can also be observed
qualitatively in the melspectrogram
in Figure \ref{fig:spectrograms}.

\emph{\textbf{Observation 4} Supervised
learning loss and MCD
are not well-correlated with the 
quality of generated speech, but language
modelling metrics like perplexity may be.} 

A common pattern across all the models
that we trained, with varying 
hyperparameter combinations,
has been that the quality of speech
monotonously increases until it plateaus. 
However, the validation loss
follows the traditional trend of
first decreasing to a minima, then
gradually increasing again before 
also reaching a plateau. This is 
surprising since an improvement
in the speech quality is not
reflected in the training 
dynamics of the model. Clearly,
this indicates a gap in the 
supervised learning objective
used for speech synthesis (mean squared error) and the true objective, which
is speech quality. We also compute
Mel Cepstral Distortion (MCD) over
ten samples generated from each of our
models and report the results 
in Table \ref{tab:results1}. Similar
to validation loss, we find MCD is not 
strongly correlated with speech quality 
either. Following from Observation 3
above, however, we find that the 
latent embeddings from the Top-line
model perform the best at predicting 
the next vowel, followed by the 
Multitask-Learning model. This indicates
that traditional text language model
metrics like perplexity and 
next-token-prediction accuracy might
be better indicators of generated speech 
quality. 
This calls for further research in
better metrics that capture 
speech quality more closely and that
can be directly optimised in training 
speech language models.




\section{Conclusion}
We present a novel speech language model
along the lines of recent work
in the textless-NLP domain. Our model
uses melspectrogram representations
and builds on top of linguistic units
like syllables or phonemes, instead
of using raw waveforms of complete
speech utterances. With several orders of
magnitude lesser data, and significantly
smaller models, our model closely
approximates babbling - 
melspectrograms of our generated speech
show clear indication of syllabic 
structure being learned. The dataset
we use is approximately 18 hours
of single-speaker read speech, and is 
several orders of magnitude lesser
than that used by \cite{Dieleman2021VariablerateDR} and
GSLM. We leave the study of scaling
laws of our speech language model
to future work.

The strength of our model also lies
in its high-impact application
areas. In speech language pathology,
for example, mis-spelling of particular
phonemes and syllables are indicative
of particular speech and language
impairments. Sub-word language models
such as ours can be used to build
effective speech recognition and
synthesis systems that can be used
as diagnostic tools and learning aids 
for affected individuals \cite{Dudy2018AutomaticAO, Dinh2020ImprovingSI}. Our method
is also useful for low resource
languages. Specifically, similar to 
\cite{Tomokiyo2005ForeignAI},
the use of syllables allows to 
generate accented speech in a 
different language, such as,
using a Telugu dataset to generate
Telugu-sounding Marathi speech.

Through our analyses we show the
capability of our different models
in capturing information in their
latent representations that is 
useful to
language-modelling. Most notable
is the observation that Panphon
representations, which have previously
proved successful in improving
performance of text based natural
language tasks, also show promise in
the speech language modelling domain.
Finally, we highlight an important
research gap in speech quality 
assessment and show that while
validation loss and MCD are not 
strongly correlated with MOS for our
models, traditional text language modelling
metrics like perplexity and 
next-token-prediction accuracy could be.

\vfill\pagebreak


\bibliographystyle{IEEEbib}
\bibliography{refs}
\section{Appendix}

\begin{table}[h!]
\centering
    \small
\begin{tabular}{l c c}
\toprule
Model &  MOS & MCD \\
\midrule
\multicolumn{3}{c}{Syllable Level}\\
\midrule
Synthesis-only & 3.3 & 3.15 \\
MTL (Panphon) & 3.4 & 3.14 \\
Auxiliary (LSTM-LM) & 2.8 & 3.08 \\
Auxiliary (RoBERTa) & 2.9 & 3.10 \\
Auxiliary (Top-line) & 3.7 & 3.13 \\
\bottomrule
\end{tabular}
\caption{Mean Opinion Score (MOS) and Mean Mel Cepstral Distortion (MCD) over 10 speech samples generated from our 5 different models. \label{tab:results1}}
\end{table}

\end{document}